\documentclass[11pt,a4paper]{article}
\usepackage[hyperref]{acl2019}
\usepackage{times}
\usepackage{latexsym}
\usepackage{graphicx}
\usepackage{url}
\usepackage{amsmath}
\usepackage{bm}
\usepackage{amssymb}
\usepackage{CJK}
\usepackage{booktabs}
\usepackage{tabu}

\aclfinalcopy 
\def\aclpaperid{2670} 

\newcommand*\myat{{\fontfamily{ptm}\selectfont\small @}}

\title{Modeling Semantic Relationship in Multi-turn Conversations with Hierarchical Latent Variables}

\def\first{$^1$}
\def\second{$^2$}
\def\third{$^3$}
\def\comma{$^,$}
\def\star{$^*$}

\author{Lei Shen\first\comma\second ~~~ Yang Feng\first\comma\second\star
~~~ Haolan Zhan\second\comma\third
\\
{ \first {Key Laboratory of Intelligent Information Processing,}} \\ Institute of Computing Technology, Chinese Academy of Sciences, Beijing, China   \\
{ \second {University of Chinese Academy of Sciences, Beijing, China}} \\
{ \third {State Key Laboratory of Computer Science,}} \\ Institute of Software, Chinese Academy of Sciences, Beijing, China \\ 
{\tt \small \{\href{mailto:shenlei17z@ict.ac.cn}{shenlei17z}, \href{mailto:fengyang@ict.ac.cn}{fengyang}\}\myat ict.ac.cn}	\quad {\href{mailto:zhanhl@ios.ac.cn}{{\tt\small zhanhl\myat ios.ac.cn}}}
}

\date{}

\begin{document}
\maketitle
\newcommand\blfootnote[1]{%
\begingroup 
\renewcommand\thefootnote{}\footnote{#1}%
\addtocounter{footnote}{-1}%
\endgroup
}
\begin{abstract}
\blfootnote{\star{Corresponding Author}}Multi-turn conversations consist of complex semantic structures, and it is still a challenge to generate coherent and diverse responses given previous utterances. It's practical that a conversation takes place under a background, meanwhile, the query and response are usually most related and they are consistent in topic but also different in content. However, little work focuses on such hierarchical relationship among utterances. To address this problem, we propose a Conversational Semantic Relationship RNN (CSRR) model to construct the dependency explicitly. The model contains latent variables in three hierarchies. The discourse-level one captures the global background, the pair-level one stands for the common topic information between query and response, and the utterance-level ones try to represent differences in content. Experimental results show that our model significantly improves the quality of responses in terms of fluency, coherence and diversity compared to baseline methods.

\end{abstract}

\section{Introduction}



Inspired by the observation that real-world human conversations are usually multi-turn, some studies have focused on multi-turn conversations and taken context (history utterances in previous turns) into account for response generation. How to model the relationship between the response and context is essential to generate coherent and logical conversations. Currently, the researchers employ some hierarchical architectures to model the relationship. \newcite{serban2016building} use a context RNN to integrate historical information, \newcite{tian2017make} sum up all utterances weighted by the similarity score between an utterance and the query, while \newcite{zhang2018context} apply attention mechanism on history utterances. Besides, \newcite{xing2018hierarchical} add a word-level attention to capture fine-grained features. 

In practice, we usually need to understand the meaning of utterances and capture their semantic dependency, not just word-level alignments \cite{luo2018auto}. As shown in Table \ref{tab:example}, this short conversation is about speaker A asks the current situation of speaker B. At the beginning, they talk about B's position. Then in the last two utterances, both speakers think about the way for B to come back. A mentions ``{\it umbrella}'', while B wants A to ``{\it pick him/her up}''. What's more, there is no ``word-to-word'' matching in query and response. Unfortunately, the aforementioned hierarchical architectures do not model the meaning of each utterance explicitly and has to summarize the meaning of utterances on the fly during generating the response, and hence there is no guarantee that the inferred meaning is adequate to the original utterance. To address this problem, variational autoencoders (VAEs) \cite{kingma2013auto} are introduced to learn the meaning of utterances explicitly and a reconstruction loss is employed to make sure the learned meaning is faithful to the corresponding utterance. Besides, more variations are imported into utterance level to help generate more diverse responses.

\begin{table}[htb]
\centering
\begin{tabular}{l}
  \hline
  A: Where are you?\\
  B: I'm stuck in my office with rain.\\
  A: Didn't you bring your umbrella?\\
  \hline
  B: No. Please come and pick me up.\\
  \hline
\end{tabular} 
\caption{An example of the semantic relationship in a multi-turn conversation.}
\label{tab:example}
\end{table}


However, all these frameworks ignore the practical situation that a conversation usually takes place under a background with two speakers communicating interactively and query is the most relevant utterance to the response. Hence we need to pay more attention to the relationship between query and response. To generate a coherent and engaging conversation, query and response should be consistent in topic and have some differences in content, the logical connection between which makes sure the conversation can go on smoothly.

On these grounds, we propose a novel \textbf{Conversational Semantic Relationship RNN (CSRR)} to explicitly learn the semantic dependency in multi-turn conversations. CSRR employs hierarchical latent variables based on VAEs to represent the meaning of utterances and meanwhile learns the relationship between query and response. Specifically, CSRR draws the background of the conversation with a discourse-level latent variable and then models the consistent semantics between query and response, e.g. the topic, with a common latent variable shared by the query and response pair, and finally models the specific meaning of the query and the response with a certain latent variable for each of them to capture the content difference. With these latent variables, we can learn the relationship between utterances hierarchically, especially the logical connection between the query and response. What is the most important, the latent variables are constrained to reconstruct the original utterances according to the hierarchical structure we define, making sure the semantics flow through the latent variables without any loss. Experimental results on two public datasets show that our model outperforms baseline methods in generating high-quality responses.

\section{Approach}
Given $n$ input messages $\{{\bf u}_t\}_{t=0}^{n-1}$, we consider the last one ${\bf u}_{n-1}$ as query and others as context. ${\bf u}_n$ denotes corresponding response.

\begin{figure}[htb]
\begin{center}
   \includegraphics[width=1.0\linewidth]{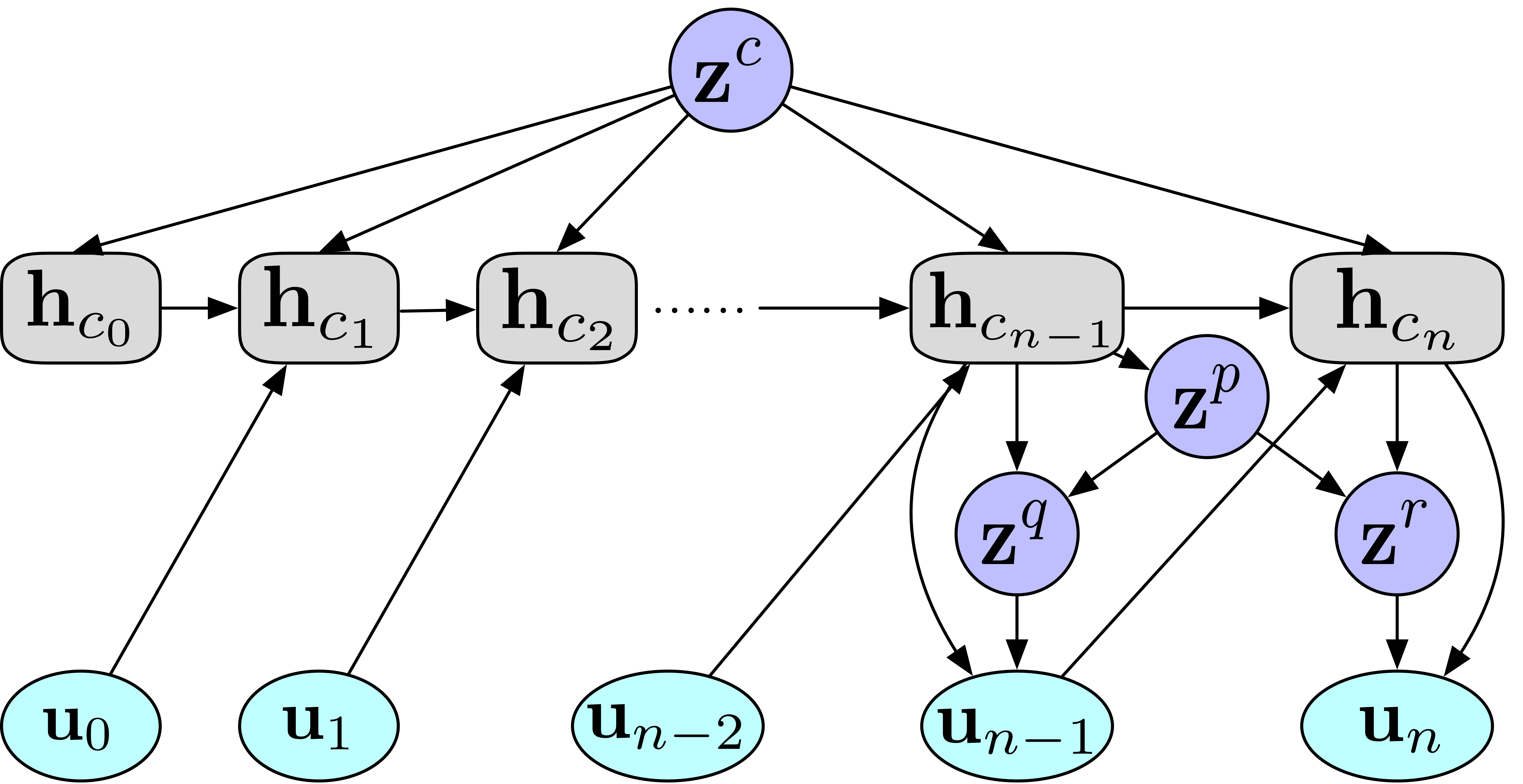}
\end{center}
   \caption{Graphical model of CSRR. ${\bf u}_t$ is the $t$-th utterance, ${\bf h}_{c_t}$ encodes context information up to time $t$.}
\label{fig:model}
\end{figure}

The proposed model is shown in Figure \ref{fig:model}. We add latent variables in three hierarchies to HRED \cite{serban2016building}. ${\bf z}^c$ is used to control the whole background in which the conversation takes place, ${\bf z}^p$ is for the consistency of topic between query and response pair, ${\bf z}^q$ and ${\bf z}^r$ try to model the content difference in each of them, respectively. For simplicity of equation description, we use $n-1$ and $n$ as the substitution of $q$ and $r$.

\subsection{Context Representation}
Each utterance ${\bf u}_t$ is encoded into a vector ${\bf v}_t$ by a bidirectional GRU (BiGRU), $f_\theta^{utt}$:
\begin{equation}
    {\bf v}_t = f_\theta^{utt}({\bf u}_t)
\end{equation}
For the inter-utterance representation, we follow the way proposed by \citet{park2018hierarchical}, which is calculated as:
\begin{equation}
    {\bf h}_{c_t} =\left\{
    \begin{aligned}
    & \text{MLP}_\theta({\bf z}^c),&  &\text{if $t$ = 0}  \\
    & f_\theta^{ctx}({\bf h}_{c_{t-1}},{\bf v}_{t-1}, {\bf z}^c),&  &\text {otherwise}
    \end{aligned}
    \right.
\end{equation}
$f_\theta^{ctx}(\cdot)$ is the activation function of GRU. ${\bf z}^c$ is the discourse-level latent variable with a standard Gaussian distribution as its prior distribution, that is:
\begin{equation}
    p_\theta({\bf z}^c) = \mathcal{N}({\bf z|0,I})
\end{equation}
For the inference of ${\bf z}^c$, we use a BiGRU $f^c$ to run over all utterance vectors $\{{\bf v}_t\}_{t=0}^n$ in training set. ($\{{\bf v}_t\}_{t=0}^{n-1}$ in test set):
\begin{align}
    q_\phi({\bf z}^c|& {\bf v}_0, ..., {\bf v}_n) = \mathcal{N}({\bf z}|\bm{\mu}^c, \bm{\sigma}^c{\bf I}) \\
    \text{where}~ & {\bf v}^c = f^c({\bf v}_0, ..., {\bf v}_n) \\
    & \bm{\mu}^c = \text{MLP}_\phi({\bf v}^c) \\
    & \bm{\sigma}^c = \text{Softplus}(\text{MLP}_\phi({\bf v}^c))
\end{align}
MLP($\cdot$) is a feed-forward network, and Softplus function is a smooth approximation to the ReLU function and can be used to ensure positiveness \cite{park2018hierarchical, serban2017hierarchical, chung2015recurrent}.

\subsection{Query-Response Relationship Modeling}
According to VAEs, texts can be generated from latent variables \cite{shen2017conditional}. Motivated by this, we add two kinds of latent variables: pair-level and also utterance-level ones for query and response.

As depicted in Figure \ref{fig:model}, ${\bf h}_{{\bf c}_{n-1}}$ encodes all context information from utterance ${\bf u}_0$ to ${\bf u}_{n-2}$. We use ${\bf z}^p$ to model the topic in query and response pair. Under the same topic, there are always some differences in content between query and response, which is represented by ${\bf z}^q$ and ${\bf z}^r$, respectively. We first define the prior distribution of ${\bf z}^p$ as follows:
\begin{equation}
    p_\theta({\bf z}^p|{\bf u}_{<n-1}, {\bf z}^c) = \mathcal{N}({\bf z}|\bm{\mu}_{n-1}, \bm{\sigma}_{n-1}{\bf I})
\end{equation}
${\bf u}_{<n-1}$ denotes utterances $\{{\bf u}_i\}_{i=0}^{n-2}$, $\bm{\mu}_{n-1}$ and $\bm{\sigma}_{n-1}$ are calculated as:
\begin{align}
    &\bm{\mu}_{n-1} = \text{MLP}_\theta({\bf h}_{{\bf c}_{n-1}}, {\bf z}^c) \\
    &\bm{\sigma}_{n-1} = \text{Softplus}(\text{MLP}_\theta({\bf h}_{{\bf c}_{n-1}}, {\bf z}^c)) 
\end{align}
Since ${\bf z}^q({\bf z}^{n-1})$ and ${\bf z}^r({\bf z}^n)$ are also under the control of ${\bf z}^p$, we define the prior distributions of them as:
\begin{equation}
    p_\theta({\bf z}^i|{\bf u}_{<i}, {\bf z}^c, {\bf z}^p) = \mathcal{N}({\bf z}|\bm{\mu}_i, \bm{\sigma}_i{\bf I})
\end{equation}
Here, $i = n-1$ or $n$. The means and the diagonal variances are computed as:
\begin{align}
    &\bm{\mu}_i = \text{MLP}_\theta({\bf h}_{{\bf c}_i}, {\bf z}^c, {\bf z}^p) \\
    &\bm{\sigma}_i = \text{Softplus}(\text{MLP}_\theta({\bf h}_{{\bf c}_i}, {\bf z}^c, {\bf z}^p))
\end{align}
The posterior distributions are:
\begin{align}
    & q_\phi({\bf z}^p|{\bf u}_{\leq{n-1}}, {\bf z}^c) = \mathcal{N}({\bf z}|\bm{\mu}_{n-1}^{'}, \bm{\sigma}_{n-1}^{'}{\bf I}) \\
    & q_\phi({\bf z}^i|{\bf u}_{\leq{i}}, {\bf z}^c, {\bf z}^p) = \mathcal{N}({\bf z}|\bm{\mu}_i^{'}, \bm{\sigma}_i^{'}{\bf I})
\end{align}
$q_\phi(\cdot)$ is a recognition model used to approximate the intractable true posterior distribution. The means and the diagonal variances are defined as:
\begin{align}
    &\bm{\mu}_{n-1}^{'} = \text{MLP}_\phi({\bf v}_{n-1}, {\bf v}_n, {\bf h}_{{\bf c}_{n-1}}, {\bf z}^c) \label{eq:16}\\
    &\bm{\sigma}_{n-1}^{'} = \text{Softplus}(\text{MLP}_\phi({\bf v}_{n-1}, {\bf v}_n, {\bf h}_{{\bf c}_{n-1}}, {\bf z}^c)) \label{eq:17}\\
    &\bm{\mu}_i^{'} = \text{MLP}_\phi({\bf v}_i, {\bf h}_{c_i}, {\bf z}^c, {\bf z}^p) \label{eq:18}\\
    &\bm{\sigma}_i^{'} = \text{Softplus}(\text{MLP}_\phi({\bf v}_i, {\bf h}_{c_i}, {\bf z}^c, {\bf z}^p)) \label{eq:19}
\end{align}
Note that in Equation \ref{eq:16} and \ref{eq:17}, both ${\bf v}_{n-1}$ and ${\bf v}_n$ are taken into consideration, while Equation \ref{eq:18} and \ref{eq:19} use ${\bf z}^p$ and corresponding ${\bf v}_i$.

\subsection{Training}
Because of the existence of latent variables in query-response pair, we use decoder $f_\theta^{dec}$ to generate ${\bf u}_{n-1}$ and ${\bf u}_n$:
\begin{equation}
    p_\theta({\bf u}_i|{\bf u}_{<i}) = f_\theta^{dec}({\bf u}_i|{\bf h}_{{\bf c}_i}, {\bf z}^c, {\bf z}^p, {\bf z}^i)
\end{equation}

The training objective is to maximize the following variational lower-bound:
\begin{equation}
\begin{aligned}
    &\log p_\theta({\bf u}_{n-1}, {\bf u}_n | {\bf u}_0, ..., {\bf u}_{n-2}) \geq \\
    & \mathbb{E}_{q_\phi}[\log p_\theta({\bf u}_i|{\bf z}^c, {\bf z}^p, {\bf z}^i, {\bf u}_{<i})] \\
    &-D_{KL}(q_\phi({\bf z}^c|{\bf u}_{\leq n})||p_\theta({\bf z}^c)) \\
    &-D_{KL}(q_\phi({\bf z}^p|{\bf u}_{\leq n})||p_\theta({\bf z}^p|{\bf u}_{<{n-1}})) \\
    &-\sum_{i=n-1}^n D_{KL}(q_\phi({\bf z}^i|{\bf u}_{\leq i})||p_\theta({\bf z}^i|{\bf u}_{<i})) 
\label{eq:obj}
\end{aligned}
\end{equation}
Equation \ref{eq:obj} consists of two parts: the reconstruction term and KL divergence terms based on three kinds of latent variables.

\section{Experiment}
\subsection{Experimental Settings}
\textbf{Datasets:} We conduct our experiment on {\bf Ubuntu Dialog Corpus} \cite{lowe2015ubuntu} and {\bf Cornell Movie Dialog Corpus} \cite{danescu2011chameleons}. As Cornell Movie Dialog does not provide a separate test set, we randomly split the corpus with the ratio 8:1:1. For each dataset, we keep conversations with more than 3 utterances. The number of multi-turn conversations in train/valid/test set is 898142/19560/18920 for Ubuntu Dialog, and 36004/4501/4501 for Cornell Movie Dialog.

\noindent\textbf{Hyper-parameters:} 
In our model and all baselines, Gated Recurrent Unit (GRU) \cite{cholearning} is selected as the fundamental cell in encoder and decoder layers, and the hidden dimension is 1,000. We set the word embedding dimension to 500, and all latent variables have a dimension of 100. 
For optimization, we use Adam \cite{kingma2014adam} 
with gradient clipping. The sentence padding length is set to 15, and the max conversation length is 10. In order to alleviate degeneration problem of variational framework \cite{bowman2015generating}, we also apply KL annealing \cite{bowman2015generating} in all models with latent variables. The KL annealing steps are 15,000 for Cornell Movie Dialog and 250,000 for Ubuntu Dialog. 

\noindent\textbf{Baseline Models:}
We compare our model with three baselines. They all focus on multi-turn conversations, and the third one is a state-of-the-art variational model. 1) Hierarchical recurrent encoder-decoder (HRED) \cite{serban2016building}. 2) Variational HRED (VHRED) \cite{serban2017hierarchical} with word drop (w.d) and KL annealing \cite{bowman2015generating}, the word drop ratio equals to 0.25. 3) Variational Hierarchical Conversation RNN (VHCR) with utterance drop (u.d) \cite{park2018hierarchical} and KL annealing, the utterance drop ratio equals to 0.25.

\begin{table*}[!htb]
    \centering
    \resizebox{\linewidth}{26mm}{
    \begin{tabu}{l|ccccc|ccc}\hline
    \tabucline[1pt]
        \bf Model & \bf Average & \bf Extrema & \bf Greedy & \bf Dist-1 & \bf Dist-2 & \bf Coherence & \bf Fluency & \bf Informativeness\\ \hline
        \multicolumn{9}{c}{\bf Ubuntu Dialog}\\ \hline
        HRED & 0.570 & 0.329 & 0.415 & 0.494 & 0.814 & 2.96 & 3.64 & 2.89  \\ \hline
        VHRED+w.d & 0.556 & 0.312 & 0.405 & 0.523 & 0.856 & 2.52 & 3.35 & 3.24  \\ \hline
        VHCR+u.d & 0.572 & 0.330 & 0.416 & 0.512 & 0.837 & 2.42 & 3.48 & 2.99 \\ \hline
        CSRR & \textbf{0.612} & \textbf{0.345} & \textbf{0.457} & \textbf{0.561} & \textbf{0.882} & \textbf{3.39} & \textbf{3.91} & \textbf{3.75} \\ \hline
        \multicolumn{9}{c}{\bf Cornell Movie Dialog}\\ \hline
        HRED & 0.547 & 0.370 & 0.387 & 0.489 & 0.801 & 3.02 & 3.65 & 2.85 \\ \hline
        VHRED+w.d & 0.556 & 0.365 & 0.405 & 0.512 & 0.850 & 3.05 & 3.76 & 3.24\\ \hline
        VHCR+u.d & 0.587 & 0.378 & 0.434 & 0.507 & 0.837 & 3.13 & 3.73 & 3.06 \\ \hline
        CSRR & \textbf{0.620} & \textbf{0.395} & \textbf{0.462} & \textbf{0.522} & \textbf{0.873} & \textbf{3.43} & \textbf{3.82} & \textbf{3.78}\\[-1pt] \tabucline[1pt]
        \hline
    \end{tabu}}
    \caption{Automatic and human evaluation results on Ubuntu Dialog Corpus and Cornell Movie Dialog Corpus.}
    \label{tab:eval}
\end{table*}

\subsection{Evaluation Design}
Open-domain response generation does not have a standard criterion for automatic evaluation, like BLEU \cite{papineni2002bleu} for machine translation. Our model is designed to improve the coherence/relevance and diversity of generated responses. To measure the performance effectively, we use 5 automatic evaluation metrics along with human evaluation.

\noindent\textbf{Average, Greedy and Extrema:} Rather than calculating the token-level or n-gram similarity as the perplexity and BLEU, these three metrics are embedding-based and measure the semantic similarity between the words in the generated response and the ground truth \cite{serban2017hierarchical, liu2016not}. We use word2vec embeddings trained on the Google News Corpus \footnote{\url{https://code.google.com/archive/p/word2vec/}} in this section. Please refer to \newcite{serban2017hierarchical} for more details.

\noindent\textbf{Dist-1 and Dist-2:} Following the work of \newcite{li2016diversity}, we apply {\it Distinct} to report the degree of diversity. {\it Dist-1/2} is defined as the ratio of unique uni/bi-grams over all uni/bi-grams in generated responses.

\noindent\textbf{Human Evaluation:} 
Since automatic evaluation results may not be fully consistent with human judgements \cite{liu2016not}, human evaluation is necessary. Inspired by \newcite{luo2018auto}, we use following three criteria. \textbf{Fluency} measures whether the generated responses have grammatical errors. \textbf{Coherence} denotes the semantic consistency and relevance between a response and its context. \textbf{Informativeness} indicates whether the response is meaningful and good at word usage. A general reply should have the lowest \textit{Informativeness} score. Each of these measurement scores ranges from 1 to 5. We randomly sample 100 examples from test set and generate total 400 responses using models mentioned above. All generated responses are scored by 7 annotators, who are postgraduate students and not involved in other parts of the experiment.

\subsection{Results of Automatic Evaluation}
The left part of Table \ref{tab:eval} is about automatic evaluation on test set. The proposed CSRR model significantly outperforms other baselines on three embedding-based metrics on both datasets. The improvement of our model indicates our semantic relationship modeling better reflects the structure of real-world conversations, and the responses generated by our models are more relevant to context. As for diversity, CSRR also gets the highest Dist-1 and Dist-2 scores.

For Ubuntu Dialog dataset, VHRED+w.d is the worst. With the help of discourse-level latent variable and utterance drop, VHCR+u.d leads to better performance. However, HRED is the worst on the Cornell Movie dataset. \newcite{park2018hierarchical} empirically explained the difference based on that Cornell Movie Dialog dataset is small in size, but very diverse and complex in content and style, and models like HRED often fail to generate appropriate responses for the context.


\subsection{Results of Human Evaluation}
The right part of Table \ref{tab:eval} is about human evaluation results on 400 (100$\times$4) responses. First, it is clear that CSRR model receives the best evaluation on three aspects, which proves the effectiveness of CSRR on generating high quality responses. Second, because of the existence of discourse-level and pair-level latent variables, responses are more coherent. Since these two kinds of variables learn high level semantic information, utterance-level ones serve better on expression diversion, also improve sentence fluency and informativeness.

\begin{table}[!htb]
\centering
\small
\begin{tabular}{p{7.2cm}}
  \hline
  \tabucline[1pt]
  \textbf{Example 1}: \\
  \hline
  \textbf{Context}: how's sleeping beauty? $\mapsto$ still sleeping. $\mapsto$ kiss him, maybe he'll turn into a frog. \\
  \hline
  \textbf{HRED}: he's a good man.\\
  \textbf{VHRED+w.d}: what?\\
  \textbf{VHCR+u.d}: maybe he's dead. \\
  \textbf{CSRR}: Really? I won't try.\\
  \hline
  \tabucline[1pt]
  \textbf{Example 2}: \\
  \hline
  \textbf{Context}: tried installing the driver from nvidia, but funny thing is i've installed $\mapsto$ what is the problem you are having exactly $\mapsto$ at this point driver activated but not in use \\
  \hline
  \textbf{HRED}: what card do you have?\\
  \textbf{VHRED+w.d}: are you using the nvidia driver? or from the command line? \\
  \textbf{VHCR+u.d}: i have no idea what you are talking about, i just know that \\
  \textbf{CSRR}: you need to install the nvidia driver from the nvidia site.\\
  \hline
  \tabucline[1pt]
  \textbf{Example 3}: \\
  \hline
  \textbf{Context}: hi wondering whether to use 32 or 64 bit $\mapsto$ how much ram does the pc have? more than 2 gb you might $\mapsto$ definitely? because, i' ve heard that ubuntu recommends 32 bit and\\
  \hline 
  \textbf{HRED}: it's a bit more than a 32 bit version of a 32bit system.\\
  \textbf{VHRED+w.d}: i have no idea , but it doesn't matter. \\
  \textbf{VHCR+u.d}: i don't know, but i haven't used it much.\\
  \textbf{CSRR}: yes, but you can use 64 bit or 32 bit.\\
  \hline
  \tabucline[1pt]
  \textbf{Example 4}: \\
  \hline
  \textbf{Context}: how can i narrow down an issue i'm having? i am running $\mapsto$ check out the system logs in var logs ... in particular kern.log, syslog $\mapsto$ thank you very much. \\
  \hline 
  \textbf{HRED}: no problem.\\
  \textbf{VHRED+w.d}: i'm glad to hear that, don't know why.\\
  \textbf{VHCR+u.d}: i'm not sure what you mean by ` $<$unk$>$ '.\\
  \textbf{CSRR}: you're welcome. i don't know if that's the case.\\
  \tabucline[1pt]
  \hline 
\end{tabular} 
\caption{Examples generated by CSRR model and other baselines. The first example is from Cornell Movie Dialog, while the bottom three rows are from Ubuntu Dialog.}
\label{tab:casestudy}
\end{table}

\subsection{Case Study and Ablation Study}
Table \ref{tab:casestudy} shows the examples generated by CSRR model and other baseline models. For some easy questions, like greeting (Example 4), both HRED and CSRR perform well. In contrast, VHRED+w.d and VHCR+u.d tend to generate general and meaningless responses. For hard questions, like some technical ones (Example 1 to 3), the proposed CSRR obviously outperforms other baselines. Note that VHCR is to show the effectiveness of ${\bf z}^c$ and it can also be considered as the ablation study of CSRR to illustrate the validity of ${\bf z}^p$. From above cases, we empirically find that with the help of ${\bf z}^p$, response generated by CSRR are not only relevant and consistent to context, but also informative and meaningful.

\section{Conclusion and Future Work}
In this work, we propose a Conversational Semantic Relationship RNN model to learn the semantic dependency in multi-turn conversations. We apply hierarchical strategy to obtain context information, and add three-hierarchy latent variables to capture semantic relationship. According to automatic evaluation and human evaluation, our model significantly improves the quality of generated responses, especially in coherence, sentence fluency and language diversity. 

In the future, we will model the semantic relationship in previous turns, and also import reinforcement learning to control the process of topic changes.

\section*{Acknowledgements}
This work was supported by National Natural Science Foundation of China (NO. 61662077, NO. 61876174) and National Key R\&D Program of China (NO. YS2017YFGH001428). We sincerely thank the anonymous reviewers for their helpful and valuable suggestions.

\bibliography{acl2019}
\bibliographystyle{acl_natbib}
\end{document}